\documentclass[runningheads]{llncs}
\usepackage{url}
\usepackage{graphicx}
\usepackage{subcaption}
\usepackage{hyperref}
\usepackage{float}
\usepackage[strings]{underscore}
\usepackage[numbers]{natbib}

\begin{document}
\title{ROBO: Robust, Fully Neural Object Detection for Robot Soccer}
\author{Marton Szemenyei\inst{1} \and
Vladimir Estivill-Castro\inst{2}}
\authorrunning{Szemenyei \& Estivill-Castro}
\institute{Budapest University of Technology and Economics, Budapest, Hungary,
\email{szemenyei@iit.bme.hu} \and
Griffith University, Brisbane, QLD, Australia}
\maketitle              % typeset the header of the contribution
\begin{abstract}
Deep Learning has become exceptionally popular in the last few years due to its success in computer vision~\cite{DLClass,FRCNN,YOLO} and other fields of AI~\cite{Trans,QandA,Turing}. However, deep neural networks are computationally expensive, which limits their application in low power embedded systems, such as mobile robots. In this paper, an efficient neural network architecture is proposed for the problem of detecting relevant objects in robot soccer environments. The ROBO model's increase in efficiency is achieved by exploiting the peculiarities of the environment. Compared to the state-of-the-art Tiny YOLO model, the proposed network provides approximately 35 times decrease in run time, while achieving superior average precision, although at the cost of slightly worse localization accuracy.
% We would like to encourage you to list your keywords within
% the abstract section using the \keywords{...} command.
\end{abstract}
\section{Introduction}

Object detection is one of the fundamental tasks of computer vision, since accurately localizing and classifying relevant objects is a necessary component of scene understanding. Deep learning-based methods have achieved considerable success lately, and dominated this field in the last few years~\cite{FRCNN,YOLO,YOLO2}. While these methods can provide state-of-the-art performance, their high computational requirements limit their use in low-power embedded systems. 

In this paper, we present a fully neural object detection system for the Nao v6 robot, capable of detecting four object classes relevant to robot soccer (ball, line crossing, goalpost and robot) simultaneously. To achieve this, we propose the ROBO object detection architecture, which is based on the popular Tiny YOLOv3 model. ROBO exploits the idiosyncrasies of the robot soccer environment to provide a considerable (approximately 35 times) decrease in run time compared to the Tiny YOLOv3 method, while achieving superior accuracy.

The networks are pre-trained on a synthetic database, and subsequently fine-tuned and evaluated on a smaller real dataset. For the fine-tuning step, we propose a novel training method, called synthetic transfer learning, which allows us better convergence by only retraining the first few layers of the network. During fine-tuning, the networks are pruned using L1 regularization for further improvement in speed.

We evaluate three slightly different versions of the ROBO architecture on our database demonstrating their superior speed and accuracy compared with the Tiny YOLOv3 model. We also demonstrate the effects of pruning and synthetic transfer learning. Our code for training, the databases used, and several pre-trained models are available online~\cite{Github}.

\section{Previous Work}

Deep neural networks have become a widely-used and powerful solution to numerous machine learning problems. The renewed interest in DNNs was largely sparked by the dramatic increase in the availability of high quality datasets and computational resources. The lack of these was one of the major barriers to training deep neural networks, in addition to some numerical problems~\cite{BackProp}. Their applications are countless, ranging from natural language processing problems such as translation~\cite{Trans} or question-answering~\cite{QandA} to implementing Turing machines~\cite{Turing}. The use of deep learning is perhaps most prominent in the field of computer vision, where deep neural networks are used for standard classification~\cite{DLClass} and object detection~\cite{FRCNN}. Lately, the applications expanded to several more complex areas, such as image captioning~\cite{Caption}, translation and generation~\cite{GAN}.

\subsection{Object Detection}

Deep learning detection methods can be divided into two categories: region-based methods, such as Faster-RCNN~\cite{FRCNN}, and single-shot detectors, such as YOLO~\cite{YOLO} or SSD~\cite{SSD}. Region-based methods first create object proposals using a region proposal network (RPN), followed by a classification step determining the type of object in the region. These methods usually have higher run time, and can be extended for instance segmentation~\cite{MaskRCNN}.

One of the important additions of Faster-RCNN was the introduction of anchor boxes. Anchor boxes are basically template for the object bounding boxes, derived from the training dataset using clustering. During the region proposal step and the prediction of the final bounding box, the box parameters are estimated relatively to the anchor box. This solution helps the numeric convergence of these methods greatly, and was adopted by single-shot detectors as well~\cite{YOLO2}.

Single-shot detectors skip the region proposal step, and predict the bounding box and class of the objects directly from the feature map. The YOLO~\cite{YOLO2} architecture for instance, makes $k$ (the number of anchor boxes) object predictions for each cell in the downscaled feature map using a $1x1$ convolutional layer. In order to avoid false detections, every prediction includes an \emph{objectness} score signaling whether there is an actual object at that location. This means, that from every cell $k*(5+N_{class})$ numbers are predicted. Every object must be predicted by the cell that contains its center, while the bounding box is predicted relative to the cell coordinates and the $k^{th}$ anchor box. To avoid multiple detections, non maximum-suppression is employed.

YOLO is a considerably faster, if less accurate object detection method, which tends to struggle with accurately predicting small objects. To resolve this, the v3 version of the architecture uses upscaling layers and makes predictions at multiple scales~\cite{YOLO3}, similarly to the SSD method. YOLO also includes a smaller version, called Tiny YOLO~\cite{YOLO2,YOLO3}, which uses a medium-sized convolutional base network, allowing it to run real-time on some embedded GPU platforms.

\subsection{Computer Vision in Robot Soccer}

The stated goal of RoboCup Soccer is to design a team of autonomous robots that are able to defeat the world champion soccer team using FIFA rules by the middle of the 21st century. Achieving human-level vision and scene understanding is an essential component of achieving this goal. In accordance with this insight, the RoboCup environment has steadily changed from featuring objects that are easy to recognize using low-level features, such as color, to ones that greatly resemble the actual objects used in human soccer.

The vision pipelines used by the competing teams have changed in tandem, going from human-engineered vision methods~\cite{RoboVis1, RoboVis2} to pipelines relying increasingly on machine learning. Several teams have used convolutional neural networks either for binary classification tasks~\cite{RoboNet1, RoboNet2} or to detect several relevant object categories~\cite{RoboNet3,RoboNet4}. \citet{VisualMesh} used a technique called Visual Mesh to improve the performance of neural networks at multiple scales. These methods, however, use CNNs for classification only, therefore they still require a separate object proposal method, and the quality of the system may largely depend on the efficiency of the algorithm used to generate candidates for classification. A further disadvantage is that running the same neural network on potentially overlapping image regions is wasteful, since the same features are computed twice.

One of the most important advances of recent years is the work published by Hess et al.~\cite{RoboNet5} in which they present a high-quality virtual RoboCup environment created in Unreal Engine. Using their work allows one to easily create large datasets of realistic images of a soccer field along with pixel-level semantic labeling. Since the performance of a trained neural network is highly dependent on the quality and quantity of the training data, and creating a large hand-labeled database is highly time-consuming, their work was profoundly valuable for our research. Notably, \citet{SemSeg2} used a deep neural network to perform semantic segmentation on limited hardware. Their solution is capable of detecting balls and goalposts at multiple resolutions in real-time.

Last year, we proposed the first fully neural general scene understanding system~\cite{SemSeg}. This neural network was a semantic segmentation model that could detect the same four relevant categories, while running at 2-3 fps on the Nao v5 robot. By using optical flow to propagate labels between images the frame rate of the detection system could be increased to 12 fps. Also, the network was pruned~\cite{Prune} after fine-tuning to decrease the number of computations required. However, the method of weight pruning did not use the advanced methods~\cite{NVPrune} to determine which weights should be pruned.

This solution had two major disadvantages: First, the decoder part of a semantic segmentation network effectively doubles the number of computations required, making this solution unnecessarily slow. Second, due to the pixel-wise classification loss, the network struggles to detect small objects, reducing the distance at which it can detect the ball.

\section{The ROBO Architecture}

The proposed ROBO model is heavily influenced by the popular Tiny YOLO~\cite{YOLO2} architecture. Tiny YOLO is a fully convolutional network with a stride of $32$, meaning that if given a standard $416x416$ input image, the output activation array has spatial dimensions of $13x13$. The final layer of the model is a $1x1$ convolutional layer predicting $B$ bounding boxes at every location (grid cell). Each bounding box has $5+N_c$ parameters, which are the center coordinates, width and height of the bounding box, the confidence score, and the $N_c$ class scores.

Tiny YOLO also uses so-called anchor boxes, reference bounding boxes, aquired by running clustering on the boxes in the training set. The width and height of the bounding boxes are predicted relatively to one of the $B$ anchor boxes. The center coordinates are predicted relatively to the grid cell. Predicting a certain object is the responsibility of the output with the most similar anchor box at the grid cell that contains the center of the object.

\subsection{Improving the model efficiency}

Despite its name, Tiny YOLO is a medium-sized network, with 20 layers (including pooling), some of which have 512 or even 1024 channels. This network was designed to perform well on complex datasets, like the Pascal VOC or COCO. To use such a large network for object detection in the robot soccer setting would be an overkill. Therefore, we propose the ROBO architecture, which is a 16 layer, fully convolutional network with the deepest convolutional layer having only 256 channels. This reduction in the number of channels is justified, since the robot soccer environment is considerably less complex and less varied than generic object recognition.

The ROBO architecture also replaces the max pooling layers in Tiny YOLO with strided convolution. This is due to the fact that max pooling discards spatial information, which reduces the performance of neural networks even for tasks where the spatial information is considerably less important, such as classification~\cite{Caps}. Arguably the effect of using max pooling is even worse for detection. To further allow the network to preserve some spatial information during downscaling, every strided convolutional layer increases the number of channels. Furthermore, this increases the complexity of the learnable features, since ROBO's network base has 15 subsequent convolutions, while Tiny YOLO only has 9 and 11 for its two outputs respectively.

To further increase the model's speed, the input image is downscaled aggressively. The first three layers of the network are all strided convolution, thus the spatial dimensions of the feature maps are reduced eightfold by the time the first conventional convolutional layer is applied. The total stride of the network (the ratio between the spatial size of input and output) is increased from 32 to 64, making the final part of the network four times faster. Notably, the aggressive downscaling and increased stride should in theory decrease the network's accuracy for smaller objects, the replacement of max pooling should counter that to some extent.

The ROBO architecture also exploits the fixed aspect ratio of the Nao robot's camera. All variants of YOLO are prepared for images of all size and shape, requiring a complex preprocessing step, where all images are padded and resized to $416x416$. This is rather wasteful, however, since a fair share of the computation is wasted on padded parts of the image. On the other hand, the Nao robots have a 4:3 ratio camera, meaning that we can choose a fixed resolution of $k*64*(4x3)$, where $k$ is a positive integer to ensure that no pixel information or computation is wasted. In this paper, we chose $k=2$, which results in an input resolution of $512x384$.

\subsection{Exploiting the environment}

Finally, the ROBO architecture also exploits prior knowledge about the relevant objects and their arrangement to make simplifications to the final layer of the model. The algorithm exploits the following properties of soccer fields: 

\begin{itemize}
	\item There are four classes (ball, line crossing, robot, goalpost) 
	\item There is a limited number of all classes in the field 
	\item Objects of the same class are not cluttered (robots are mild exceptions to this rule)
	\item Objects of the same class have similarly shaped bounding boxes (robots are mild exceptions to this rule, since they might fall over)
\end{itemize}

For the above reason, the ROBO model uses class-specific anchor boxes, meaning that from each cell on the final $8x6$ grid, it makes exactly one prediction for each class. The anchor boxes are also computed separately per class by simply averaging the bounding box widths and heights (our method removes the need for clustering). Since the index of the anchor box now determines the class, the classification scores can be removed, meaning that our network has $N_{class}*5 = 20$ outputs. This change simplifies both the loss function and the inference process somewhat, since the classification loss no longer has to be calculated, and non-maximum suppression is no longer necessary during training. 

Since robots are mild exceptions to some of the rules above, we experimented with allocating $2$ anchor boxes for the robot class, instead of just one. However, this did not yield noticeably different results, as even the single-box version was able to detect robot objects accurately. In the cluttered-robots scenario the model often predicts a single bounding box that encompasses both robots. In our opinion, correcting this minor issue is not worth the added complexity.

Finally, we changed the output logic of Tiny YOLO slightly: instead of upscaling the final feature layer of the network to produce an upscaled output, we simply produce the upscaled output from an earlier layer in the network. Also, instead of predicting all four classes at both outputs, the original output is responsible for predicting robots and goalposts, while the upscaled one predicts balls and crossings only. Our primary reason for doing this is that the ball and crossing classes are usually much smaller than the other two, so predicting them from a higher resolution feature map will increase the localization accuracy considerably. 

\begin{figure}[htb]
	\centering
	\includegraphics[width=0.99\textwidth]{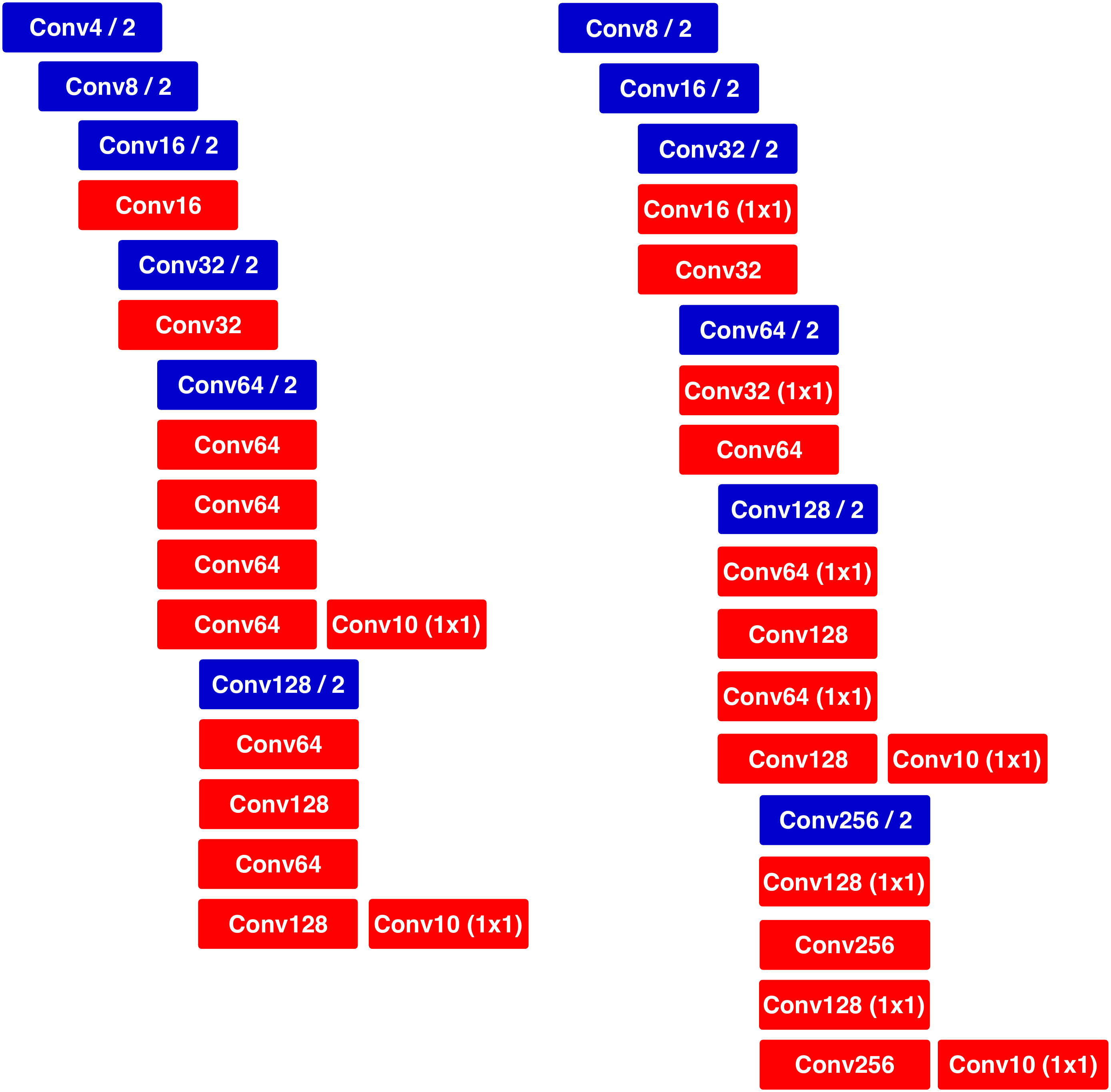}
	\caption[ The ROBO (left) and the ROBO-BN (right) architectures ]
	{\small  The ROBO (left) and the ROBO-BN (right) architectures} 
	\label{fig:net}
\end{figure}

For our experiments, we also made a slightly different version of ROBO, called ROBO-Bottleneck, or ROBO-BN for short. In this architecture, we doubled the number of channels in every convolutional layer. To account for the higher number of parameters and computational cost, we added $1\times 1$ bottleneck convolutional layers to reduce the number of channels before $3\times 3$ convolutions. The ROBO and ROBO-BN architectures are shown in Figure~\ref{fig:net}. 

\section{Training}

Our training procedure consists of two phases: First, the neural network is pre-trained on a large synthetic dataset, then the network is fine-tuned on a smaller dataset consisting of real images. The synthetic dataset was created using the Unreal Engine project published by Hess et. al.~\cite{RoboNet5} First, we created 5000 test images, using 500 scene parameter (carpet color, lighting, color temperature, etc.) variations, and 10 different scene arrangements for each. Then, we created a test set with 1250 images using 250 scene parameter sets and 5 arrangements per set. Both the parameter sets and the arrangements were generated randomly. Annotations were generated for each image automatically using the object label map generated by the engine. Bounding boxes below a certain size threshold were discarded.

A smaller real dataset was also created, with training and test sets of approximately 550 and 160 images respectively. The images were collected from four locations, including the 2017 and 2018 RoboCup venues, an outdoors venue in Melbourne, and the field set up in the MIPAL lab in Brisbane. The train and validation sets were split randomly. Both datasets contain annotations for all four relevant object classes. The images are VGA resolution, and are converted to the YUV color space.

\subsection{The training method}

Aside from the network architecture, we made a few changes to the YOLO training method as well. The training is performed using the Adam optimizer with a learning rate of $10^{-3}$. We use a cosine annealing-based learning rate schedule, with a minimal learning rate of $5*10^{-5}$. The training is ran for a total of 125 epochs. Moreover, we shuffle the training data before dividing them into minibatches of 64 images to further improve the stability of the optimization. We also use data augmentation to avoid overfitting: random horizontal flipping, and random variations of brightness, contrast, saturation and hue are applied to the images.

The most important change of our training method is the application of L1 regularization to the weights~\cite{Sparse}. While L1 regularization is primarily a way to avoid overfitting, it has a desirable side effect: sparsifying the network's weight matrices. While most state-of-the-art methods of pruning devote considerable computational expense to find the least influential weights, using L1 regularization ensures that the majority of the weights are already effectively zero and can be pruned without affecting the network at all~\cite{Sparse}. 

Still, these weights do not become exactly zero, therefore deleting them may still cause a minor disturbance. Therefore, after setting them to zero, we fine-tune the network for another 10 epochs using a learning rate of $5*10^{-5}$, while forcing the pruned weights to remain zero. The weights to be pruned are selected independently for each layer by comparing the magnitude of the weight to the largest absolute weight in the same layer. In this paper, a threshold of $0.01$ was used.

By controlling the relative weight of the regularization term, we can influence the ratio of pruned weights. By changing this weight, we trained six different versions of the same network. In the first version, no regularization was used, while in the other five we steadily increased the regularization weight, eventually achieving as much as $97\%$ on the ROBO-BN model. Note, that the weights to be pruned are not selected evenly from all layers, therefore pruning $x\%$ of the weights does not result in an execution time reduction of $x\%$.

\subsection{Synthetic transfer learning}

One of the major challenges of supervised learning is the need to use large training databases to avoid overfitting. It is well known, however, that this problems can be mitigated via transfer learning, where the neural network is first pre-trained on a large database for a different, but similar task. This allows the network to develop a basic understanding of images, which it can apply to other image-based tasks. Then, the last few layers of the network are fine-tuned on a much smaller database for the desired task. This scheme works mainly because the first part of the network performs generic feature extraction, while the last parts of the network are more task specific. By retraining the last part only, we can train the network for a different task (as long as the low level features are useful for this task as well), while the amount of free parameters is considerably less, allowing the use of a much smaller database of hand-labeled images.

Our training scheme is somewhat similar to transfer learning, in that we first pre-train the network on a large database, then we fine-tune it on a smaller one. The main difference in our case is that both databases are for the same object detection task, but come from different sources. As high quality and realistic the synthetic images may be, the distribution of their pixel values is fundamentally different from the real images. Also, the real images have complex, cluttered backgrounds, which can easily be confused with relevant foreground classes.

For the above reasons, we propose a different transfer learning scheme, in which the first few layer weights are retrained on the second database instead of the last few. We argue that this scheme is reasonable, since the first few layers of the network are responsible for extracting features from the image. We might also want to fine-tune middle-level layers to allow the network to learn more complex backgrounds. Note, that in most convolutional neural networks (including ROBO), the first few layers of the network contain much fewer parameters than the last few. This allows us to retrain more layers with similar amounts of data without overfitting, than is the standard transfer learning case.

\section{Experimental Results}
\label{sec:digErr}

We evaluated the trained networks on both datasets, computing the Mean Average Precision (mAP) of the detections. We compared the Tiny YOLO v3 network with the proposed architectures, and also examined the effect of pruning. We used three model versions: ROBO, ROBO-BN and ROBO-HR, which is a cheaper, low-resolution $(256\times 192)$ variant of ROBO. ROBO-HR is identical to ROBO, except the first strided convolutional layer is removed, ensuring that the outputs of the two models have the same scale. We determined the approximate number of operations required to run these models, and measured the average achievable FPS value of the entire vision pipeline on the Nao v5 robot using a single core.

\subsection{Comparison of detection accuracy}

We evaluated the models using several different Intersection over Union (IoU) thresholds. This threshold determines the minimum IoU value between the predicted and the ground truth bounding boxes required to consider a detection good. Lower threshold values mean that the evaluation is more lenient towards inaccurate localization. There is a slight problem with this method, however. In the case of tiny objects (such as the ball, and even more so the line crossing) even small errors in the localization can drastically decrease the IoU value. This will cause the evaluation method to be disproportionately punishing towards localization errors as opposed to classification or confidence errors.

To remedy this, we also compute the mAP values using a different error measure for localization, namely the Euclidean distance between the bounding box centers. It is worth mentioning, that this criterion ignores errors of the bounding box shape, although this is a relatively minor issue considering the rigidity of the detected objects. Table~\ref{table:1} shows the results on the synthetic and the real datasets respectively. 

\begin{table}[htb!]
	\centering
	\begin{tabular}{| c | c c c c c || c c c c c |} 
		\hline
		Database & & & Syn & & & & & Real & & \\ 
		\hline
		IoU & 0.75 & 0.5 & 0.25 & 0.1 & 0.05  & 0.75 & 0.5 & 0.25 & 0.1 & 0.05 \\ 
		\hline
		Tiny YOLO & \bf{29} & \bf{65} & 71 & 70 & 70 & \bf{21} & \bf{65} & 69 & 65 & 64 \\ 
		ROBO & 16 & 51 & 72 & 80 & 82 & 16 & 55 & 76 & 80 & \bf{82} \\ 
		ROBO-BN & 24 & 59 & \bf{77} & \bf{84} & \bf{85} & 20 & 62 & \bf{79} & \bf{81} & \bf{82} \\ 
		ROBO-HR & 15 & 48 & 69 & 77 & 78 & 16 & 52 & 73 & 79 & 80 \\ 
		\hline \hline
		Distance (px) & 4 & 8 & 16 & 32 & 64 & 4 & 8 & 16 & 32 & 64 \\ 
		\hline
		Tiny YOLO v3 & \bf{60} & 70 & 70 & 70 & 70 & \bf{41} & \bf{64} & 69 & 65 & 64 \\ 
		ROBO & 43 & 73 & 86 & 89 & 89 & 27 & 58 & \bf{80} & \bf{84} & \bf{85} \\
		ROBO-BN & 53 & \bf{79} & \bf{88} & \bf{90} & \bf{90} & 30 & 61 & \bf{80} & 83 & 83 \\ 
		ROBO-HR & 38 & 69 & 84 & 87 & 87 & 21 & 56 & 76 & 82 & 83 \\
		\hline
	\end{tabular}
	\caption{mAP comparison on the synthetic and real databases}
	\label{table:1}
\end{table}

In all four cases the ROBO architectures respond to the change in the strictness of the localization criterion much more drastically, suggesting that they struggle more with accurate localization, while Tiny YOLO struggles with accurate detection and classification. This is underscored by the fact, that using a localization criterion that is loose enough, ROBO-based models invariably manage to outperform Tiny YOLO. We believe that this is largely due to the difference between the output generation methods employed. Importantly, ROBO outperforms Tiny YOLO more decisively on the real dataset. This is most likely due to the fact, that Tiny YOLO has several times more parameters, making it much easier to overfit on a small database. 

\begin{figure}[htb]
	\centering
	\begin{subfigure}[b]{0.23\textwidth}
		\centering
		\includegraphics[width=\textwidth]{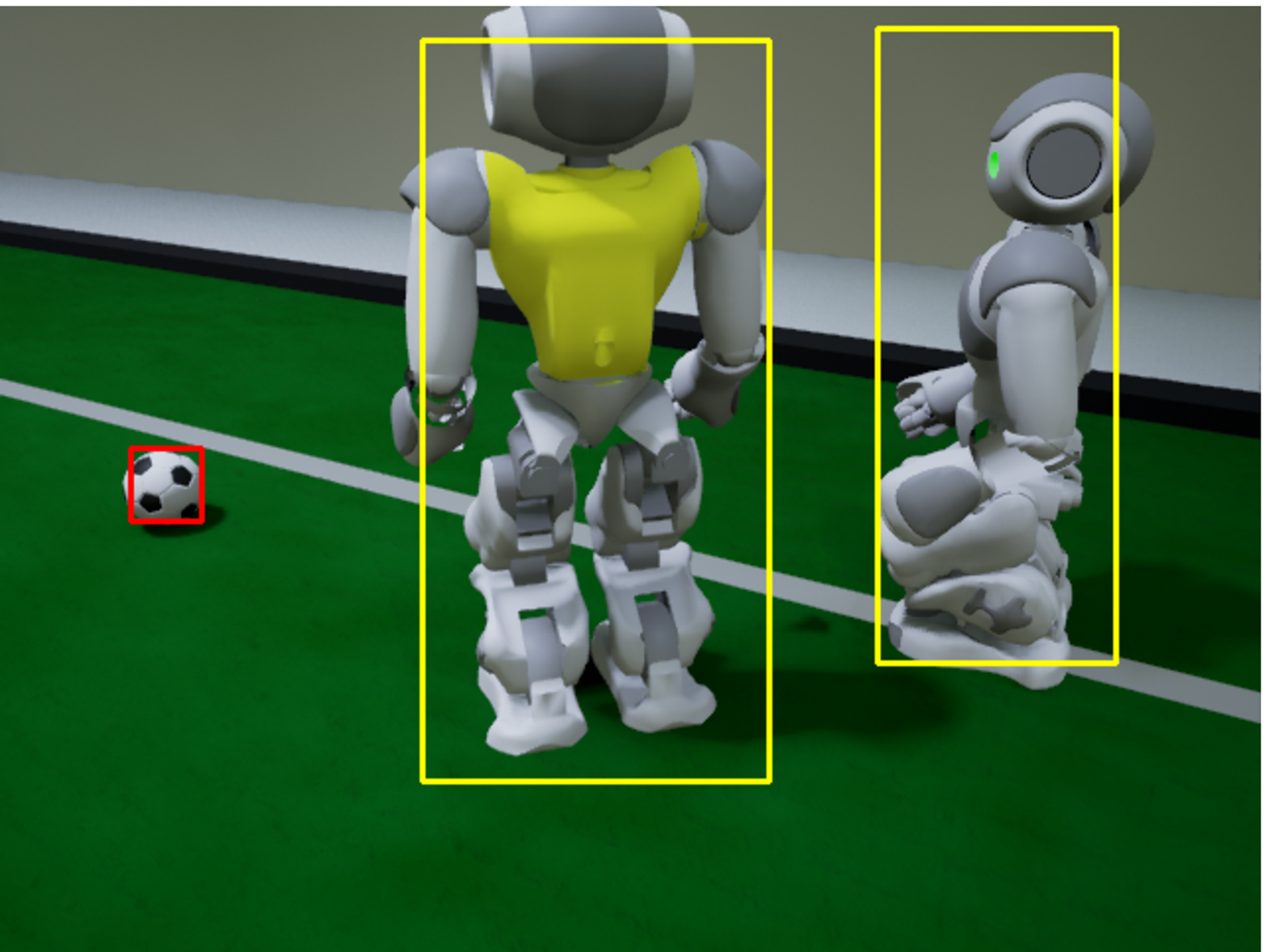}
	\end{subfigure}
	\begin{subfigure}[b]{0.23\textwidth}  
		\centering 
		\includegraphics[width=\textwidth]{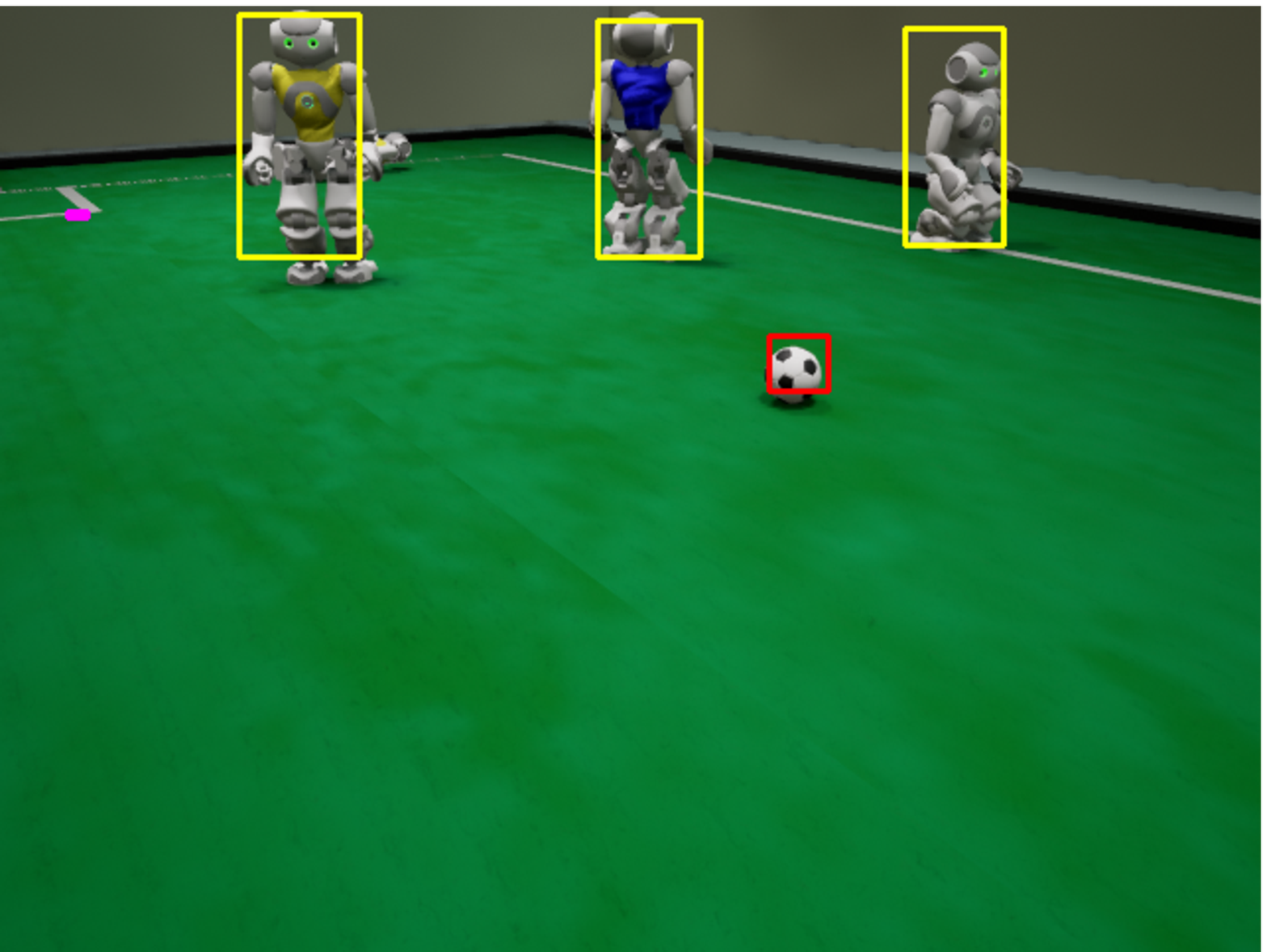}
	\end{subfigure}
	\begin{subfigure}[b]{0.23\textwidth}   
		\centering 
		\includegraphics[width=\textwidth]{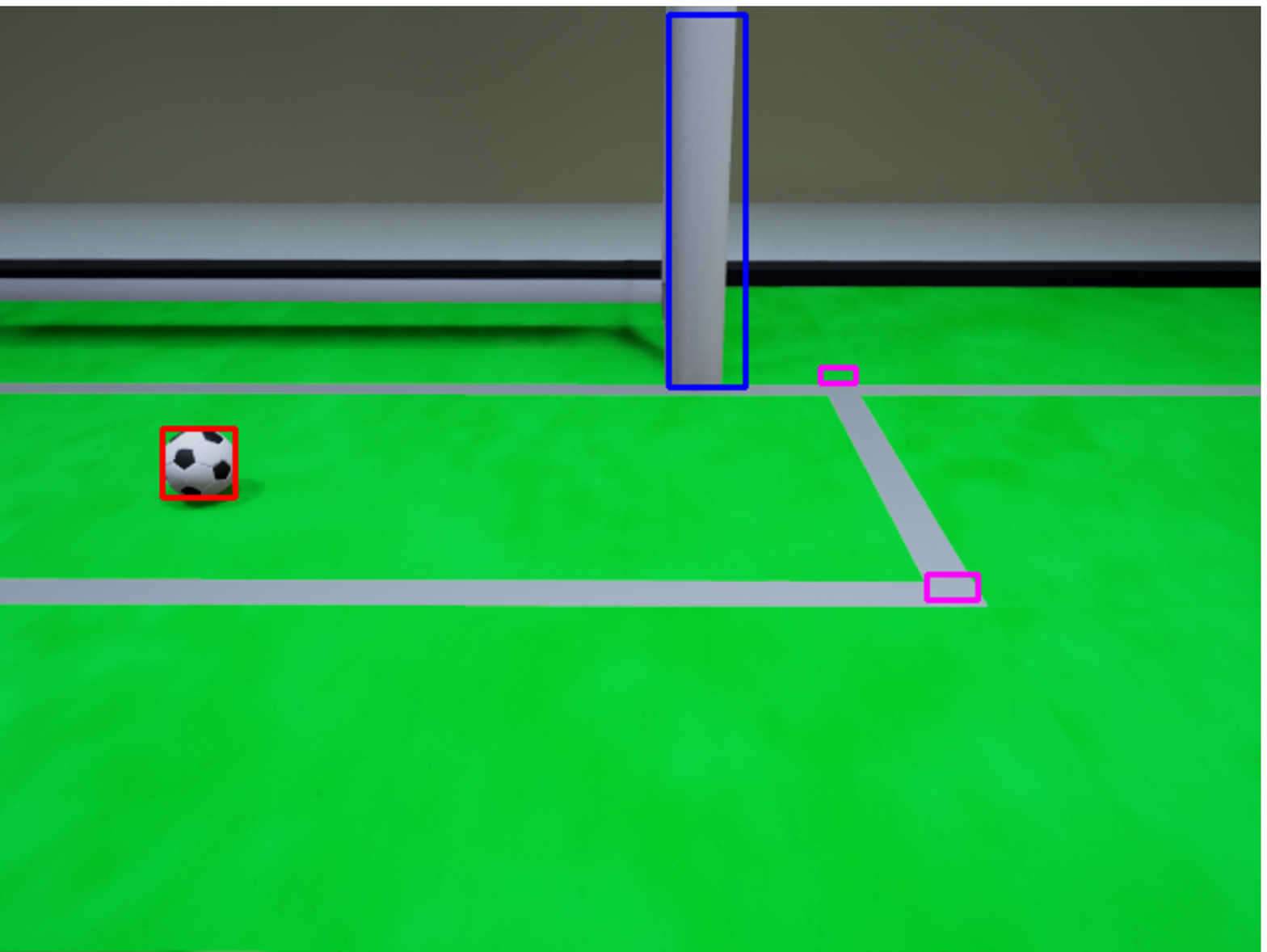}
	\end{subfigure}
	\begin{subfigure}[b]{0.23\textwidth}   
		\centering 
		\includegraphics[width=\textwidth]{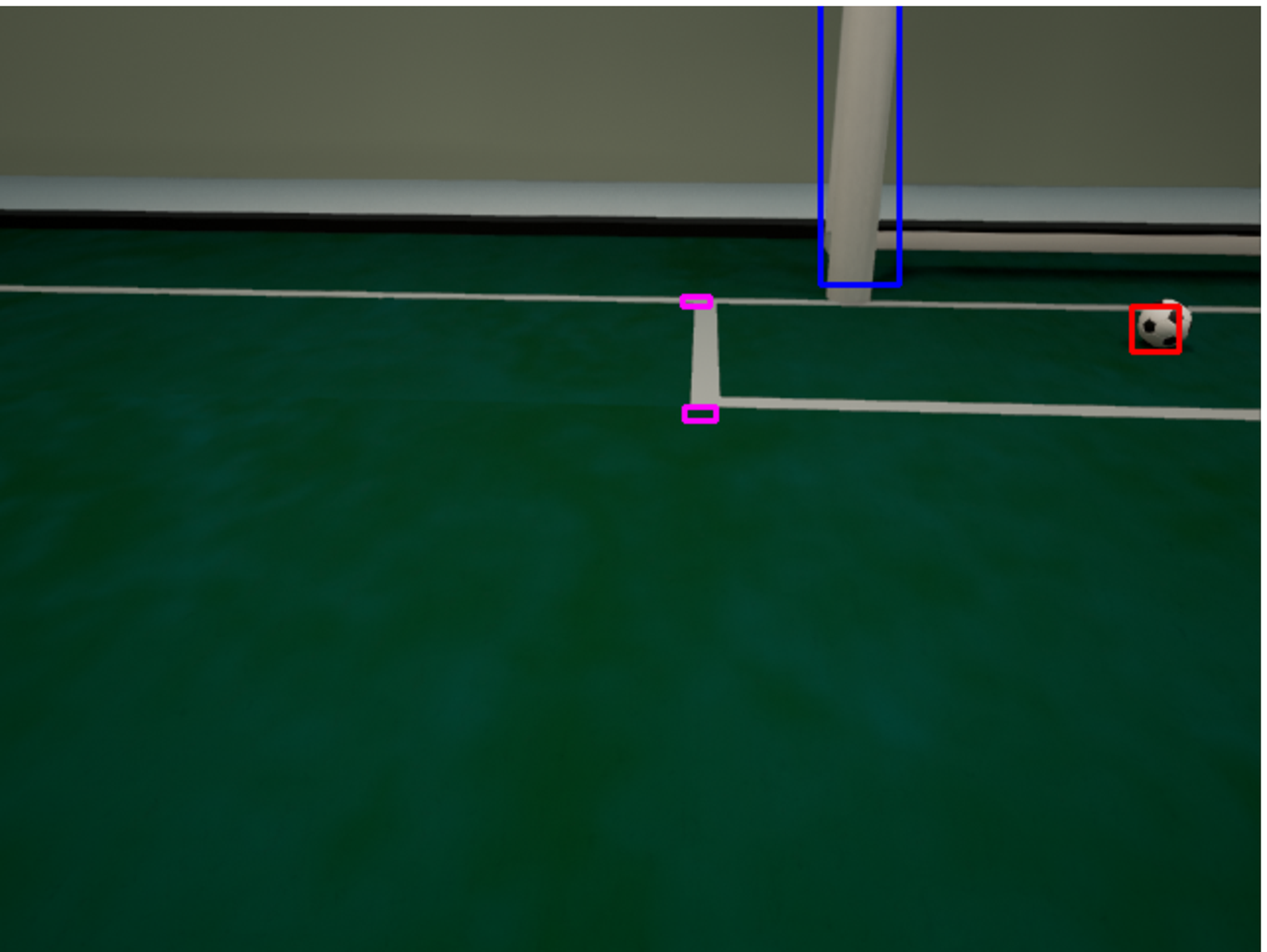}
	\end{subfigure}
	\vskip 2pt
	\begin{subfigure}[b]{0.23\textwidth}
		\centering
		\includegraphics[width=\textwidth]{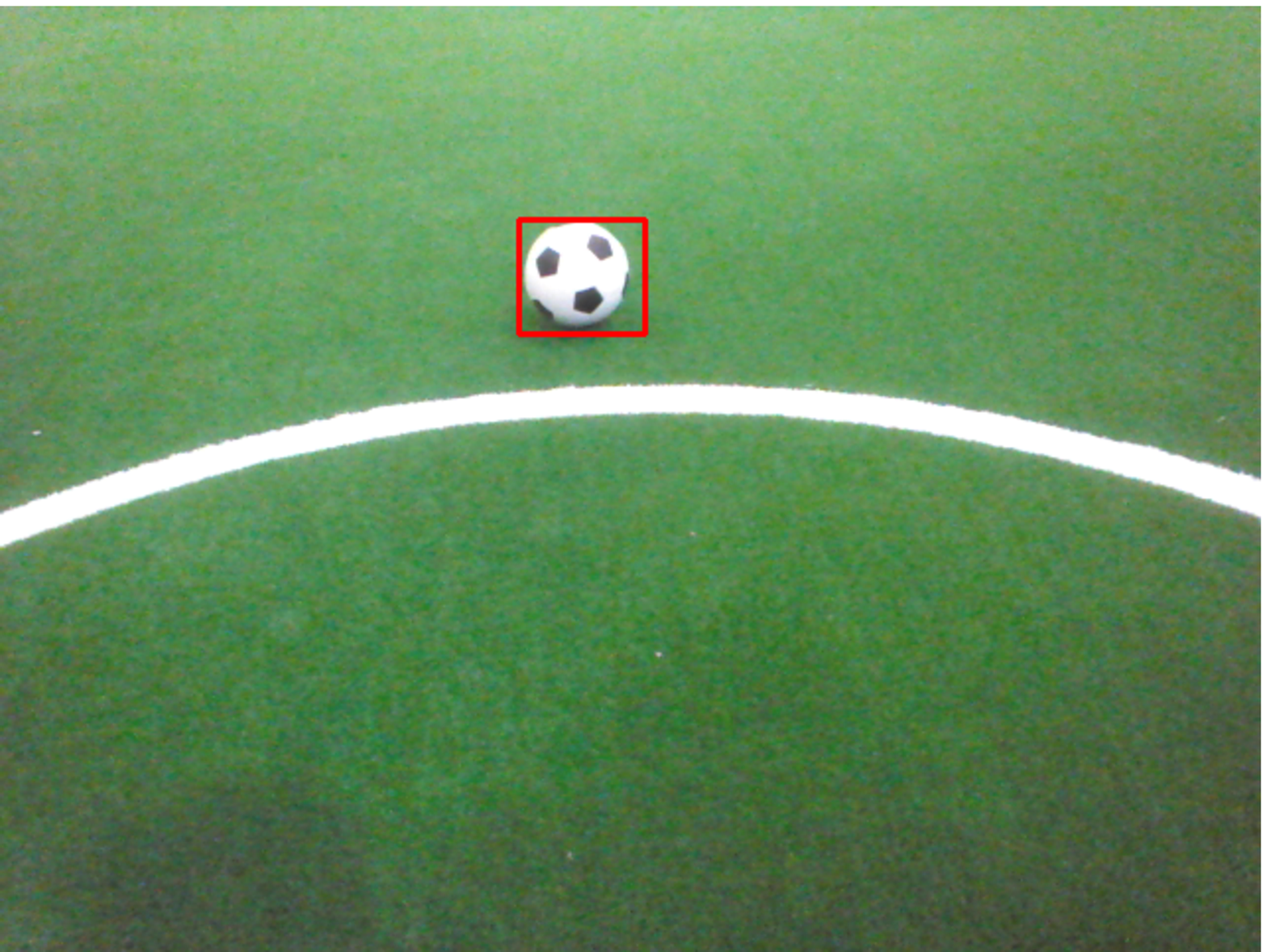}
	\end{subfigure}
	\begin{subfigure}[b]{0.23\textwidth}  
		\centering 
		\includegraphics[width=\textwidth]{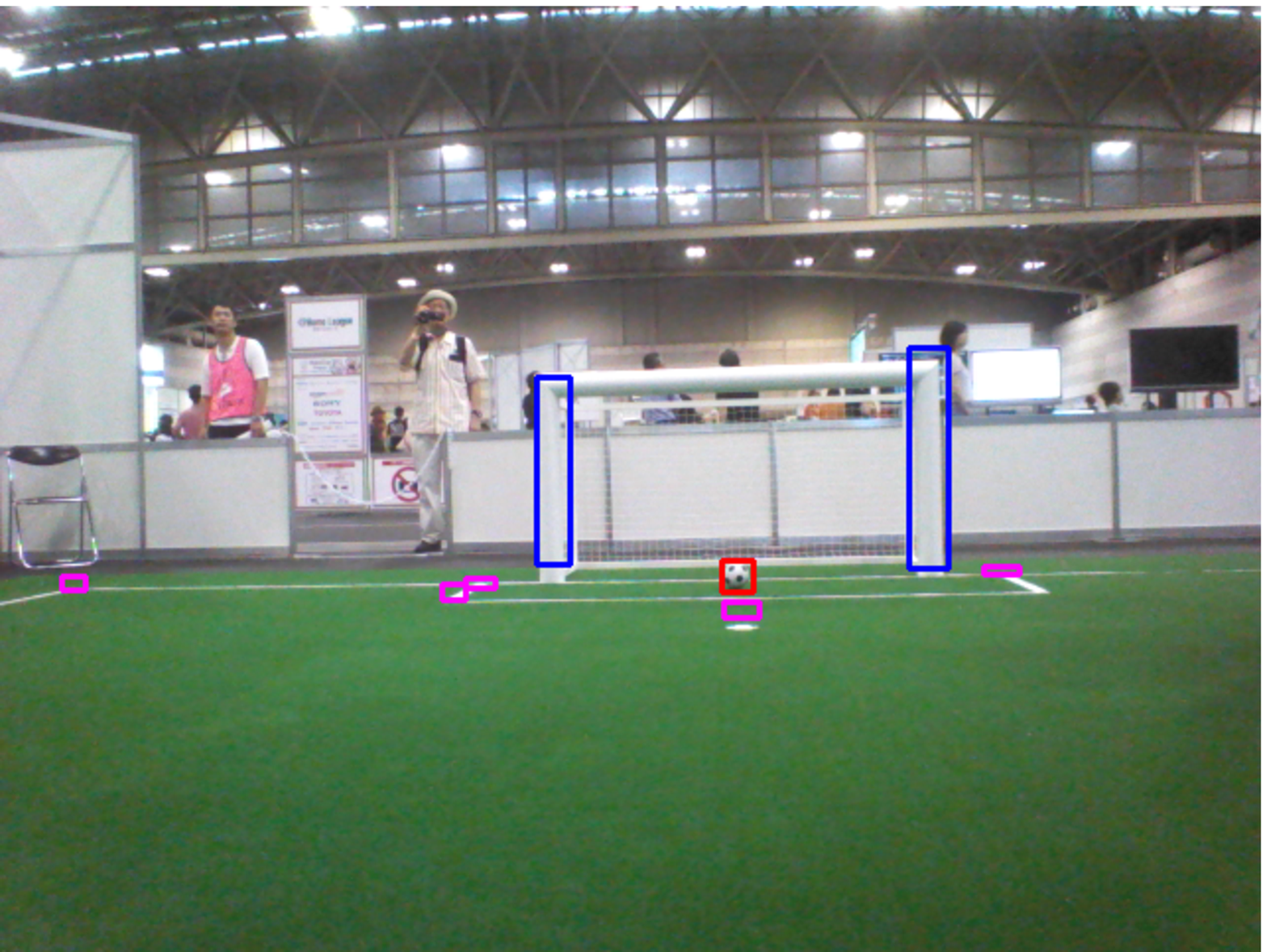}
	\end{subfigure}
	\begin{subfigure}[b]{0.23\textwidth}   
		\centering 
		\includegraphics[width=\textwidth]{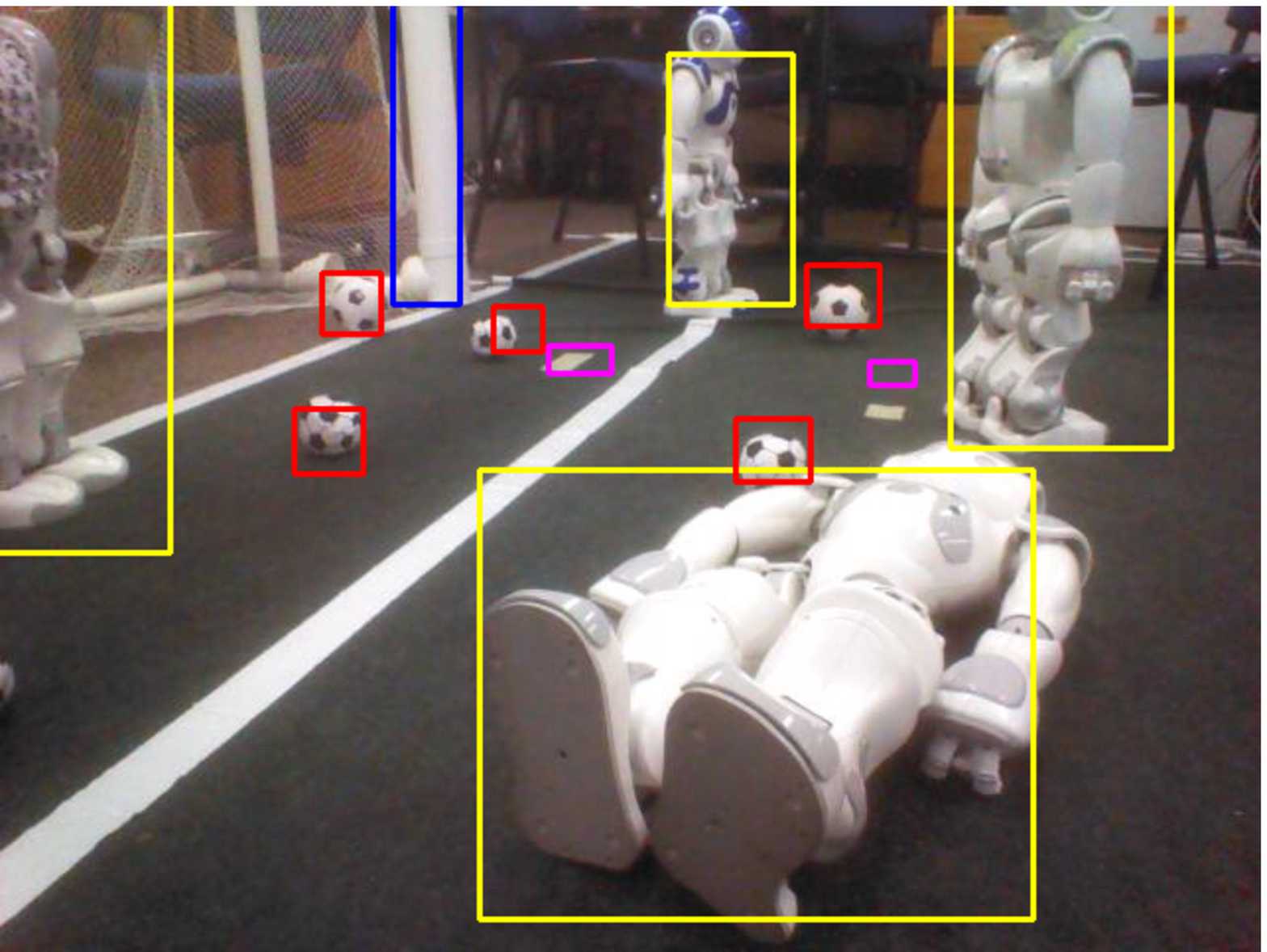}
	\end{subfigure}
	\begin{subfigure}[b]{0.23\textwidth}   
		\centering 
		\includegraphics[width=\textwidth]{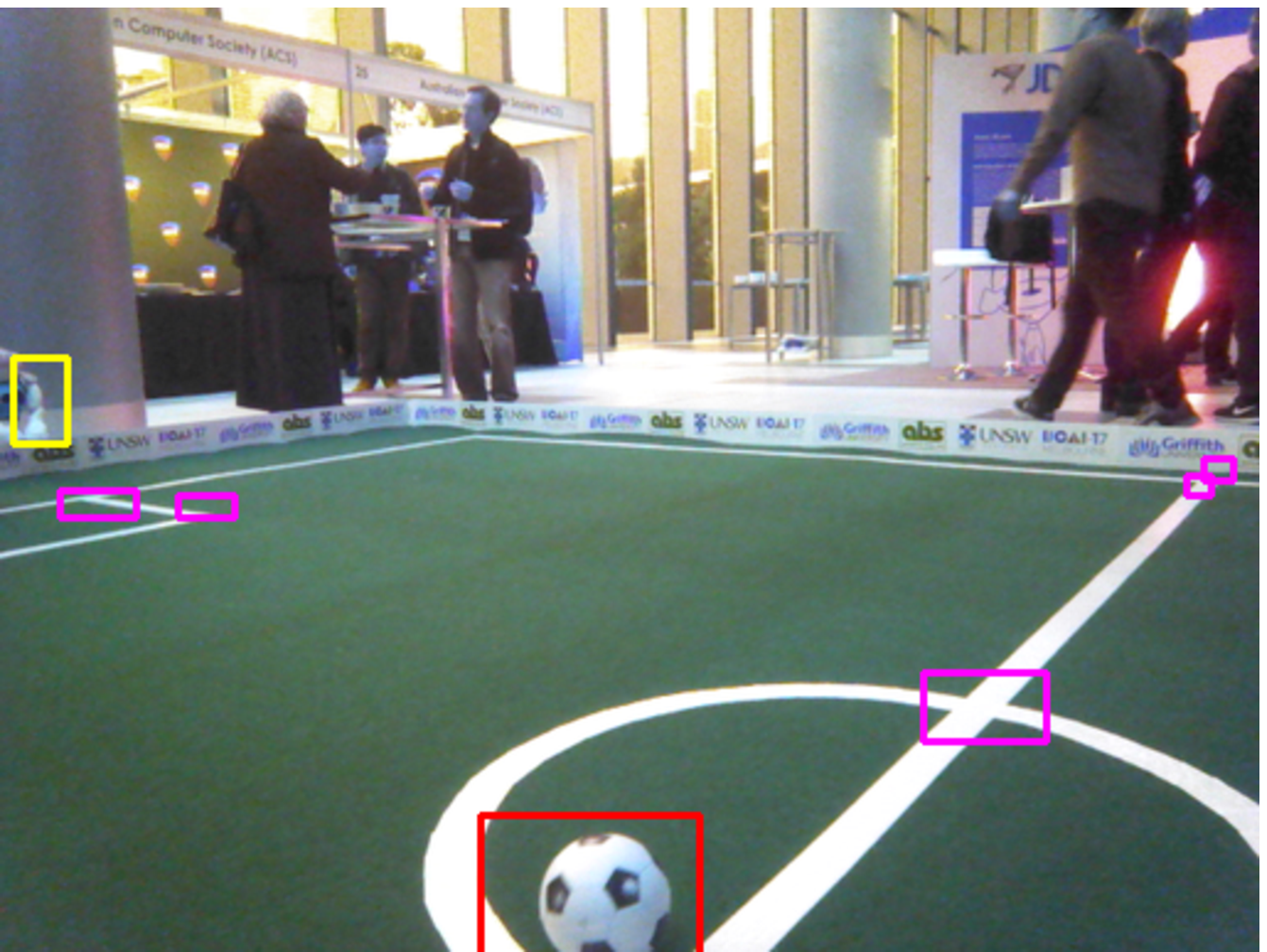}
	\end{subfigure}
	\vskip 2pt
	\begin{subfigure}[b]{0.23\textwidth}
		\centering
		\includegraphics[width=\textwidth]{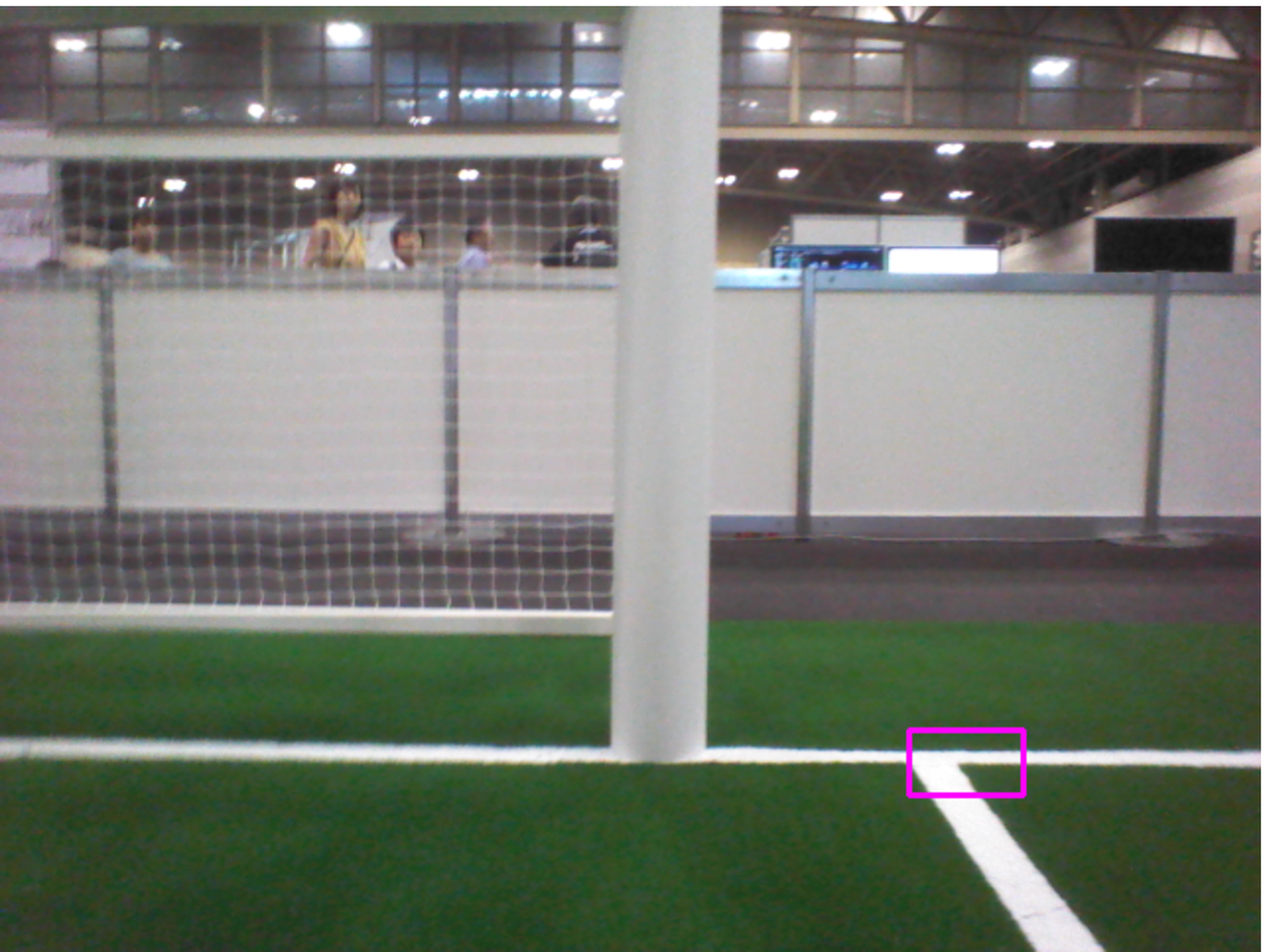}
	\end{subfigure}
	\begin{subfigure}[b]{0.23\textwidth}  
		\centering 
		\includegraphics[width=\textwidth]{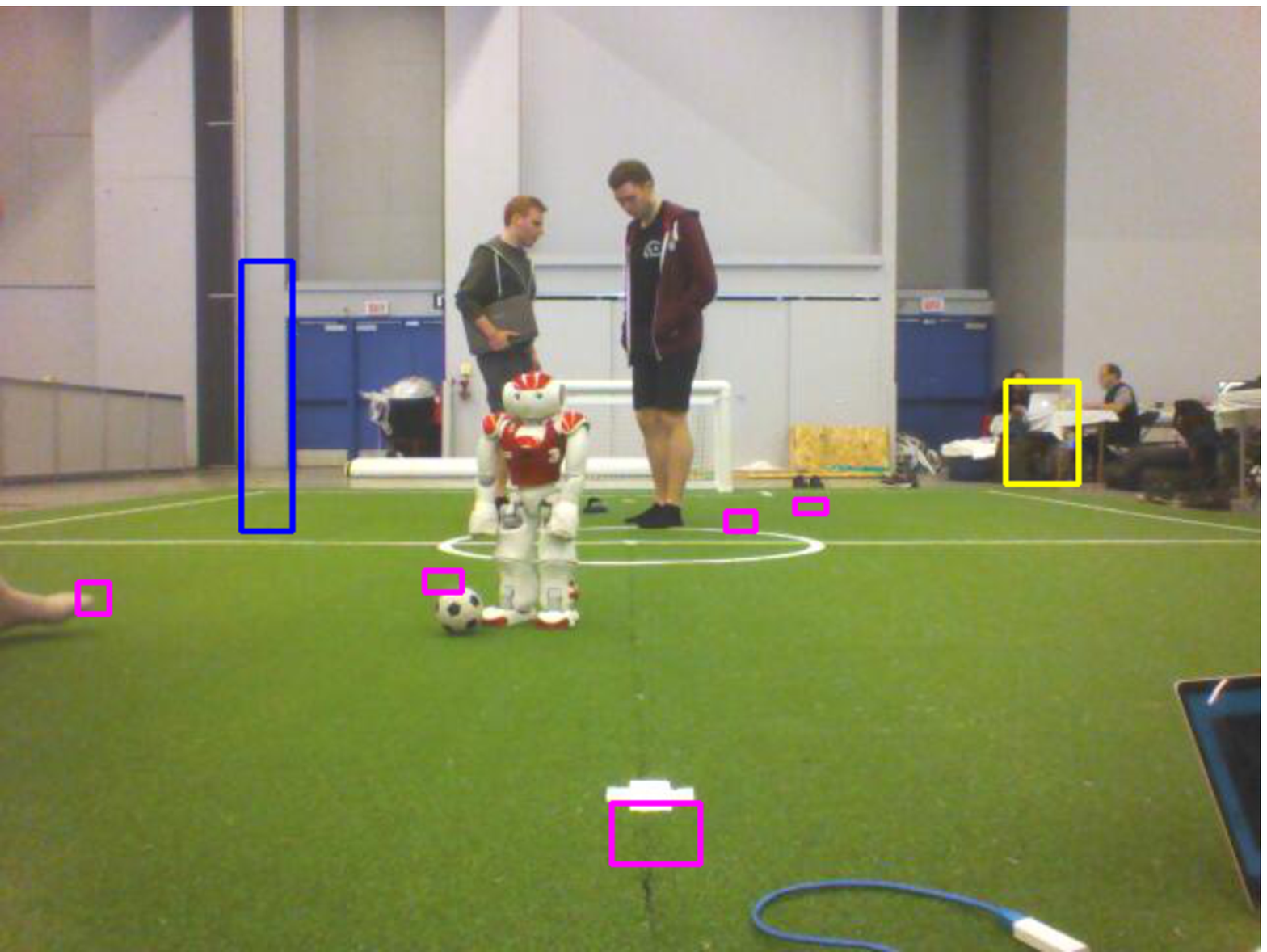}
	\end{subfigure}
	\begin{subfigure}[b]{0.23\textwidth}   
		\centering 
		\includegraphics[width=\textwidth]{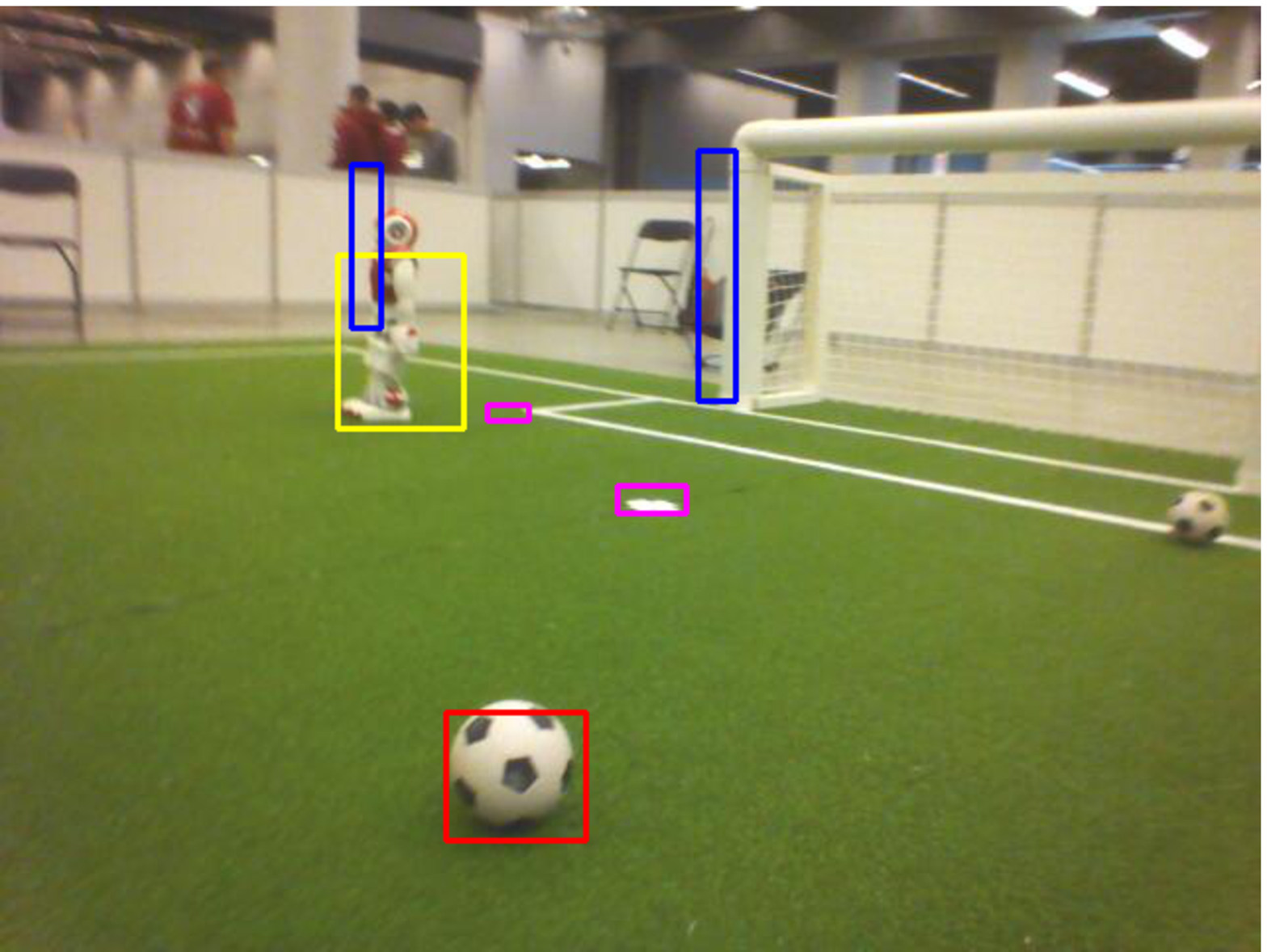}
	\end{subfigure}
	\begin{subfigure}[b]{0.23\textwidth}   
		\centering 
		\includegraphics[width=\textwidth]{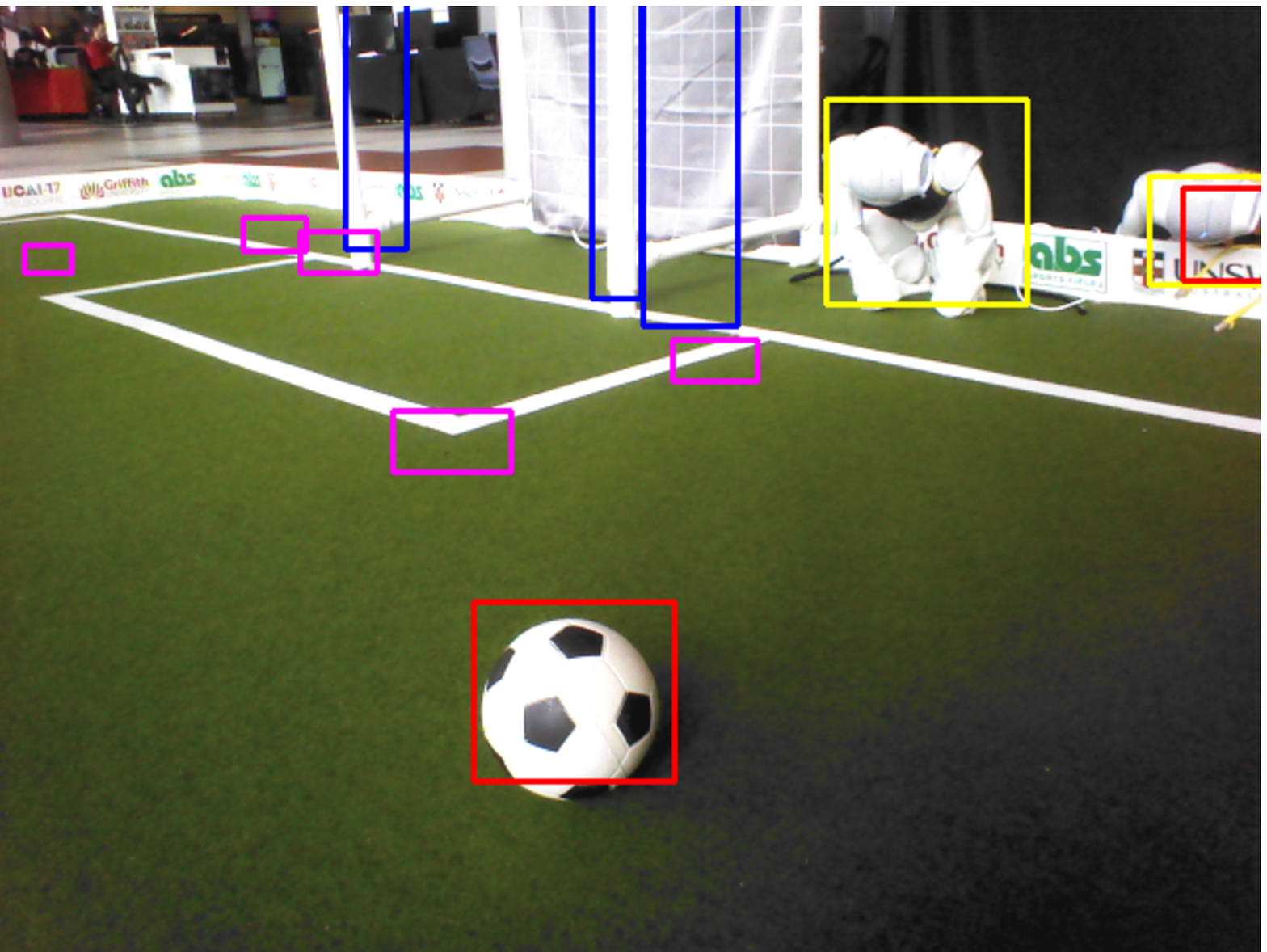}
	\end{subfigure}
	\caption[ Example results on the synthetic database (top row). Some good (middle) and bad (bottom) detection results on the real database both in- and outdoors. ]
	{\small Example results on the synthetic database (top row). Some good (middle) and bad (bottom) detection results on the real database both in- and outdoors.} 
	\label{fig:real}
\end{figure}

Predictably, the somewhat more complex ROBO-BN outperforms the other two methods, although ROBO manages to achieve similar performance on the synthetic dataset with loose criteria. This suggests that higher parameter numbers help with accurate localization. Also, the two-channel version falls short of the other two by a few percentages, but still manages to clearly outperform Tiny YOLO. Figure~\ref{fig:real} shows some example results of ROBO on the synthetic test dataset, and also some good and bad results on the real test dataset.

The per-class AP results show that the network's performance does not depend strongly on the size of the objects. The model achieves the highest AP on the ball and goalpost classes ($91$ and $87$ respectively), while it struggles more with the crossing and robot classes ($83$ and $77$). Interestingly, the largest class appears to have the smallest average precision. As demonstrated in Fig.~\ref{fig:real} the network is able to detect objects at a fair distance, although its localization is somewhat inaccurate with small objects. Recall, that qualitatively detection is more important than accurate localization, especially for small objects, since they are far.

\subsection{Effects of pruning}

\begin{figure}[htb]
	\centering
	\begin{subfigure}[b]{0.5\textwidth}
		\centering
		\includegraphics[width=0.95\textwidth]{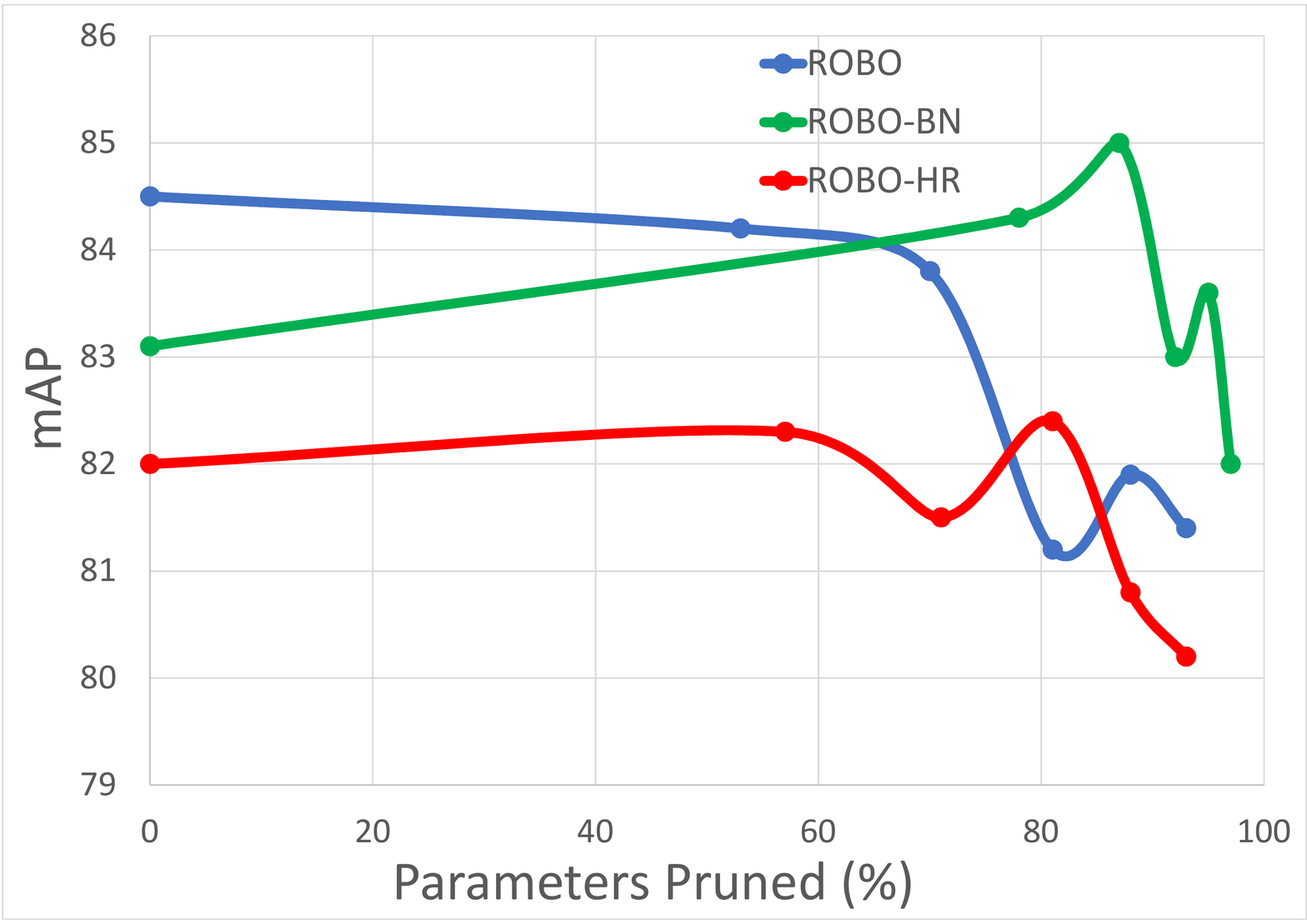}
	\end{subfigure}%
	\begin{subfigure}[b]{0.5\textwidth}  
		\centering 
		\includegraphics[width=0.95\textwidth]{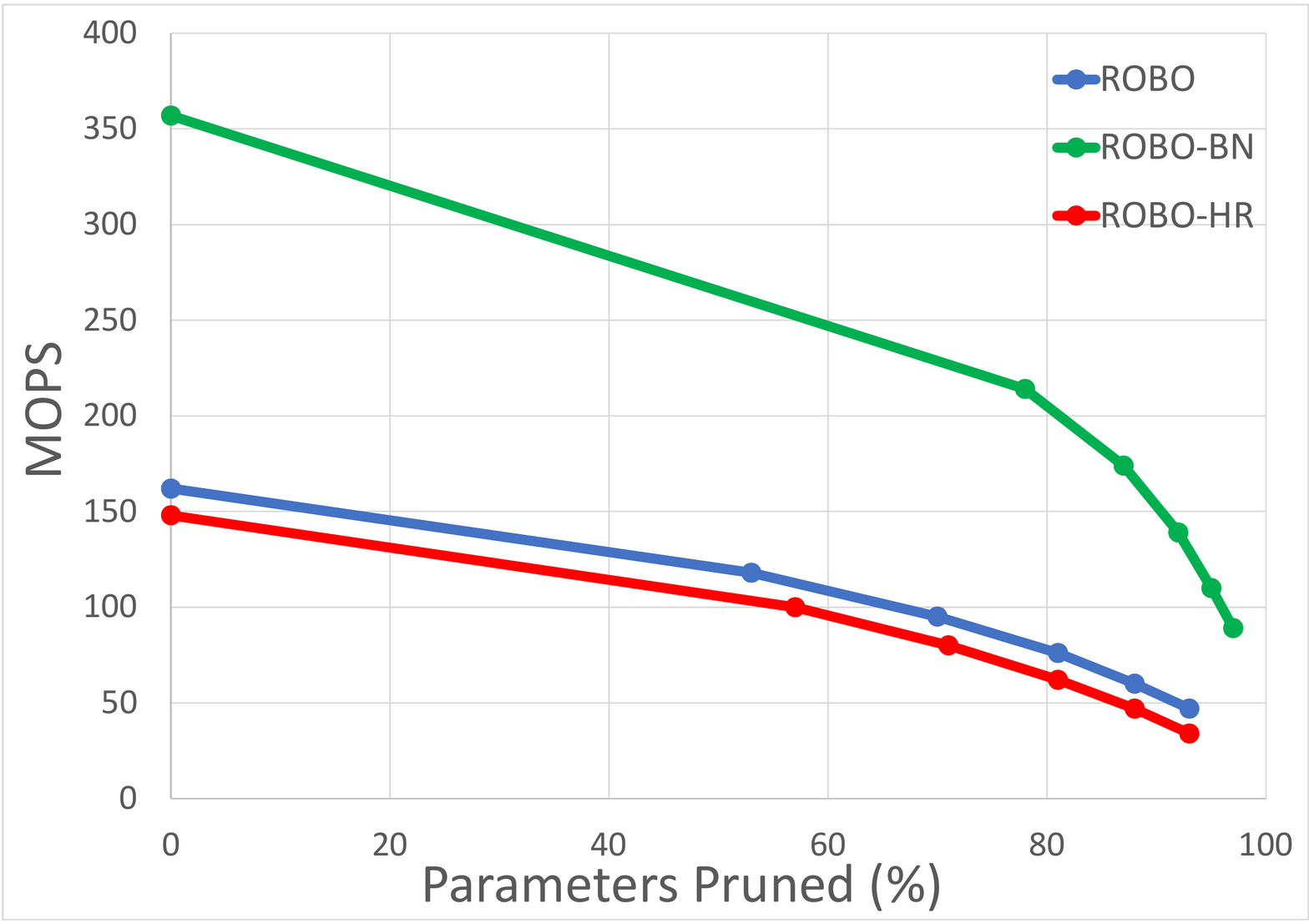}
	\end{subfigure}
	\caption[ Effect of pruning on the mAP (left) and the number of operations (right) ]
	{\small Effect of pruning on the mAP (left) and the number of operations (right)} 
	\label{fig:table3}
\end{figure}

Figure~\ref{fig:table3} shows the effect of pruning on accuracy, as well as the number of operations required for each model version. Our results show that it is possible to prune approximately $90\%$ of the ROBO model's parameters with only a negligible drop in accuracy, while reducing run time by $70\%$. Notably, the ROBO-HR version is slightly less affected by pruning, yielding a model with half the run time at the cost of a $1\%$ drop in accuracy. Also, the share of parameters pruned from ROBO-BN is higher, since it as approximately twice the number of parameters. The drop in performance is also more steep in this case, making this model slightly inferior.

The single-core run time of ROBO on the Nao v6 robot is $438$ ms, which translates to approximately $2.3$ frames per second. With $88\%$ pruning, however, the run time decreased to $179$ ms, or $5.6$ fps, while the $93\%$ version produced $146$ ms or $6.9$ fps. It is worth noting, that this is considerably slower, than expected from floating point benchmarks of the robot's processor alone. This is largely because the large initial spatial size of the images makes efficient caching more difficult. 

Moreover, using such a small network, the im2col operation that is part of the convolution takes up a considerable amount of the execution time in early layers. This motivated our ROBO-HR architecture, which performs $47\%$ fewer operations, therefore the run time on the Nao robot decreases by a similar amount to $77$ ms, which is $13$ fps. 

\subsection{Effects of synthetic transfer learning}

We also ran experiments with synthetic transfer learning, as shown in Table~\ref{table:4}. We ran several tests, where we changed the number of initial layers to retrain. By increasing this number, we allow the network to learn the real dataset better, resulting in faster convergence, but makes the network more likely to overfit. In these experiments, we fine-tuned the rest of the layers using a smaller learning rate (by a factor of ten) instead of freezing them completely. 

\begin{table}[htb!]
	\centering
	\begin{tabular}{| c | c c c c c c|} 
		\hline
		& & \multicolumn{4}{c}{ROBO} & \\
		\hline
		Layers & 0 & 3 & 5 & 7 & 9 & All   \\ 
		Parameters & 0 (0\%) & 1.5k (0.3\%) & 8.5k (1.6\%) & 36k (6.5\%) & 110k (20\%) & 555k (100\%) \\ 
		mAP @ 32px & 19.6 & 83.7 & 82.9 & 83.0 & \bf{84.6} & 84.5 \\ 
		\hline
		& & \multicolumn{4}{c}{ROBO-BN} & \\
		\hline
		Layers & 0 & 3 & 5 & 8 & 11 & All   \\ 
		Parameters & 0 (0\%) & 6k (0.4\%) & 11k (0.7\%) & 51k (3.2\%) & 221k (14\%) & 1,585k (100\%) \\ 
		mAP @ 32px & 29.3 & 82.6 & 82.8 & \bf{84.3} & 83.8 & 83.1 \\ 
		\hline
	\end{tabular}
	\caption{The results of synthetic transfer learning}
	\label{table:4}
\end{table}

The results show that retraining the first $7$ or $9$ layers only can achieve superior results to retraining the entire network, despite using only $6.5$ or $19.8$ percent of the parameters respectively. Notably, synthetic transfer learning works better for the ROBO-BN architecture, where the mAP peaked at $3.2\%$ of the parameters ($51k$) retrained. Arguably this is due to its higher parameter count, given that the ROBO architecture peaked at $110k$ retrained parameters, which is in the same order of magnitude.

\section{Conclusion}

In this paper, we presented a fully deep neural network-based object detection framework capable of detecting all relevant objects in robot soccer environments. We proposed a new architecture ROBO, and showed that it outperforms Tiny YOLO both in terms of speed and accuracy. We showed that these models can be trained using synthetic transfer learning to reduce the amount of data required. Our work also produced a small database of real images other teams can use freely. Our code, the pre-trained models, training and validation datasets are published online~\cite{Github}.

Yet, there are a few possibilities to improve our model. For instance, the network could be trained to detect the edges of the soccer field and ignore objects well outside., This way we could help the network learn by reducing the interference of complex backgrounds outside the field. Another important possibility is improving the regularization method, by using group L1 regularization~\cite{Sparse}. This would allow us to prune entire convolutional filters instead of individual weights, which in turn would make the execution of the neural network considerably more efficient.

% \printbibliography
\bibliography{szemenyei}

\end{document}